\documentclass[10pt,twocolumn,letterpaper]{article}

\usepackage[final]{cvpr}

\usepackage{graphicx}
\usepackage{amsmath}
\usepackage{amssymb}
\usepackage{booktabs}
\usepackage{float}

\usepackage[pagebackref,breaklinks,colorlinks]{hyperref}

\usepackage[capitalize]{cleveref}
\crefname{section}{Sec.}{Secs.}
\Crefname{section}{Section}{Sections}
\Crefname{table}{Table}{Tables}
\crefname{table}{Tab.}{Tabs.}

\def\cvprPaperID{19}
\def\confName{CVPR}
\def\confYear{2022}

\begin{document}

\title{MAPLE-Edge: A Runtime Latency Predictor for Edge Devices }

\author{Saeejith Nair$^1$, Saad Abbasi$^1$, Alexander Wong$^{1,2}$, Mohammad Javad Shafiee$^{1,2}$\\
$^1$University of Waterloo, Waterloo, Canada\\
$^2$DarwinAI, Waterloo, Canada\\
{\tt\small \{smnair,srabbasi,a28wong,mjshafiee\}@uwaterloo.ca}
}
\maketitle

\begin{abstract}
\vspace{-0.14in}
Neural Architecture Search (NAS) has enabled automatic discovery of more efficient neural network architectures, especially for mobile and embedded vision applications. Although recent research has proposed ways of quickly estimating latency on unseen hardware devices with just a few samples, little focus has been given to the challenges of estimating latency on runtimes using optimized graphs, such as TensorRT and specifically for edge devices. As devices like NVIDIA's Jetsons get more popular in embedded computing and robotics, we observe a pressing need to more accurately estimate inference latency of neural network architectures on diverse runtimes, including highly optimized ones. In this work, we propose MAPLE-Edge, an edge device-oriented extension of MAPLE, the state-of-the-art latency predictor for general purpose hardware, where we train a regression network on architecture-latency pairs in conjunction with a hardware-runtime descriptor to effectively estimate latency on a diverse pool of edge devices. Compared to MAPLE, MAPLE-Edge can describe the runtime and target device platform using a much smaller set of CPU performance counters that are widely available on all Linux kernels, while still achieving up to +49.6\% accuracy gains against previous state-of-the-art baseline methods on optimized edge device runtimes, using just 10 measurements from an unseen target device. We also demonstrate that unlike MAPLE which performs best when trained on a pool of devices sharing a common runtime, MAPLE-Edge can effectively generalize across runtimes by applying a trick of normalizing performance counters by the operator latency, in the measured hardware-runtime descriptor. Lastly, we show that for runtimes exhibiting lower than desired accuracy, performance can be boosted by collecting additional samples from the target device, with an extra 90 samples translating to gains of nearly +40\%.

\end{abstract}

\section{Introduction}
\label{sec:intro}
Despite the success of deep learning based approaches in computer vision tasks such as object detection~\cite{redmon2018yolov3,ren2015faster}, classification~\cite{he2016deep,simonyan2014very,tan2019efficientnet}, and segmentation~\cite{he2017mask,ronneberger2015u}, it remains a challenge to manually design neural network  architectures that are both fast, efficient and at the same time accurate. The field of Neural Architecture Search (NAS)~\cite{zoph2016neural} aims to address this problem by automating the architecture discovery process. However, most of the previous state-of-the-art techniques such as DARTS or ENAS ~\cite{liu2018darts, pham2018efficient}, formulate the optimization and search process with only the accuracy constraint as the main objective function in the search process.

The explosion of consumer applications for mobile and embedded devices necessitates the importance of multi-objective NAS methods that can discover models at the pareto-optimal frontier of accuracy, and constraints such as latency, memory, or power consumption. For embedded vision applications where extra focus gets paid to minimizing inference latency, NAS methods have relied on either measuring the model latency directly on-device~\cite{tan2019mnasnet} or estimating the inference latency using look-up-tables (LUTs)~\cite{dai2019chamnet, cai2019once,wu2019fbnet,wan2020fbnetv2, cai2018proxylessnas} or latency prediction modules (LPMs)~\cite{xu2020latency,wang2020hat,dudziak2020brp,abbasi2021maple}.

Due to the increasing size of NAS search spaces, as well as the diverse number of hardware devices and execution contexts (e.g. PyTorch, TensorFlow, TensorFlow Lite, TensorRT, OpenVino, etc.), performing measurements directly on-device is computationally intractable. Mechanisms like LUTs on the other hand are much faster at estimating latency in near constant time based on the assumption that the end-to-end latency of a network can be approximated by summing up individual latencies in a layer-wise manner. However as different studies such as Dudziak {\it et al.} showed~\cite{dudziak2020brp} such LUT based approaches are not accurate and the layer-wise latencies can significantly deviate from the true latency; which means that the supposedly pareto-optimal models predicted by NAS methods may not actually lie on the pareto-optimal frontier. Instead, state-of-the-art approaches like HELP~\cite{lee2021hardware} use learning based approaches to predict the end-to-end model latency.

While these methods perform  more accurately than LUT based estimators, the study presented in nn-Meter~\cite{zhang2021nn} showed that existing LPMs have difficulty estimating latency when inference is executed on runtimes that optimize model graphs; we also further investigate this issue here in this manuscript. For example, frameworks like TensorRT (TRT) perform kernel auto-tuning and layer and tensor fusion to accelerate inference~\cite{nvidia_tensorrt_2022}. In other words, some layers and operations may get fused or executed in parallel, compared to the sequential execution order specified by the model designer. nn-Meter tackles this by creating a more fine-grained LUT approach which characterizes models at the kernel level, but their approach involves exhaustive testing and characterization of all sampled kernels which can take anywhere from 1 to 4.4 days based on the execution runtime.

Here we propose  MAPLE-Edge, an edge device-oriented extension of the  state-of-the-art latency predictor so-called MAPLE~\cite{abbasi2021maple} for general purpose hardware, designed specifically for embedded devices. We show that by characterizing the hardware device and runtime at the operator level in just minutes (not days), the proposed MAPLE-Edge algorithm can accurately estimate the end-to-end latency of neural network architectures executed on optimized, previously unseen runtimes.

MAPLE-Edge is an LPM approach based on a hardware-aware regression model that can estimate the inference latency of a deep neural network architecture on an unseen embedded target device accurately. The proposed method specifically formulates the function $f(\boldsymbol{a}, \boldsymbol{S}; \theta) \rightarrow \hat{y}$ where $\boldsymbol{a}$ is the DNN architecture encoding, $\boldsymbol{S}$ is a quantitative hardware descriptor, $\hat{y}$ is the inference latency, and $f$ is the regression model mapping architecture encoding and hardware descriptor to the inference latency \cite{abbasi2021maple}.

\begin{figure*}[t]
  \centering
  \includegraphics[width=1.0\linewidth]{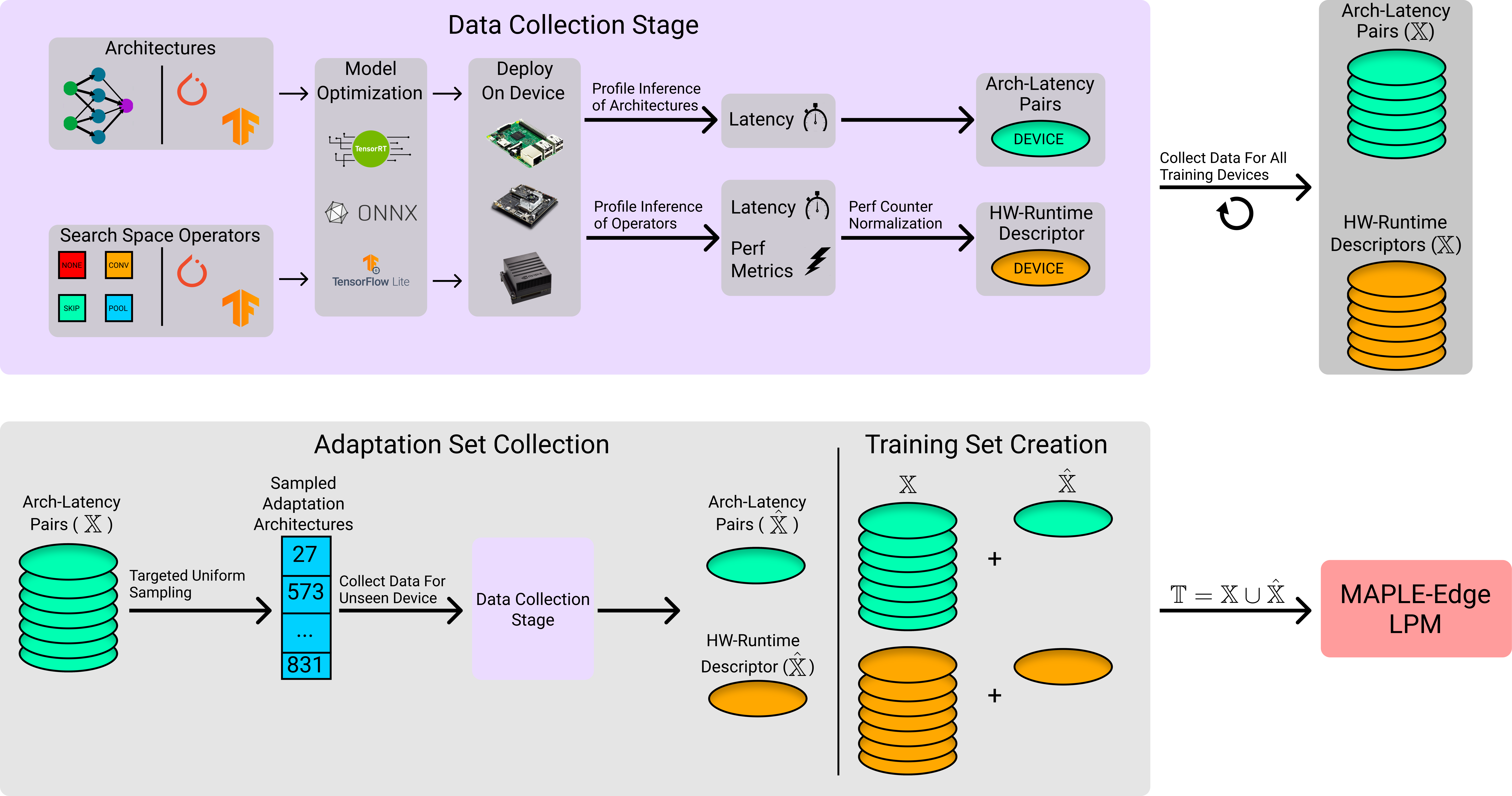}

  \caption{MAPLE-Edge overview. MAPLE-Edge is an edge device-oriented extension to MAPLE, designed specifically to estimate the latency of neural network architectures on unseen embedded devices. MAPLE-Edge trains a LPM-based hardware-aware regression model, that can effectively estimate architecture latency. To do this, it trains the LPM on a dense hardware descriptor made up of CPU performance counters, in conjunction with architecture-latency pairs. In the data collection stage, MAPLE-Edge uses an automated pipeline to convert models in NAS-Bench-201 to their optimized counterparts, deploy it to the corresponding target device, and profile inference. To adapt to previously unseen architectures, MAPLE-Edge creates a training set $\mathbb{T}$ consisting of an initial set of architectures $\mathbb{X}$ from a pool of known devices, as well as an adaptation set of architecture samples $\hat{\mathbb{X}}$ from the unseen device, which are selected through a targeted uniform sampling algorithm.}
  \label{fig:data_collection_diagram}
\end{figure*}

We show that with some simple, but important modifications to the data processing and hardware characterization stages proposed in MAPLE~\cite{abbasi2021maple}, the proposed MAPLE-Edge becomes the most effective way of estimating latency on optimized edge runtimes, paving the way forward to more accurate NAS methods for embedded vision applications. As such our key contributions are as follows:
\begin{itemize}
    \item A new latency estimator for optimized edge runtimes using an LPM algorithm is proposed. To the best of our knowledge, this is the first work    exploring the idea of using an LPM algorithm to estimate neural network architecture latency on optimized runtimes such as TensorRT. It is very important to account for the effect of runtime in latency prediction for edge devices.
    \item A new runtime characterization on embedded devices using a hardware descriptor is proposed that can be easily constructed on all Linux devices. We show that by reducing the performance metric-based hardware descriptor proposed by MAPLE into a more dense form containing performance counters that are supported by Intel and ARM processors alike, embedded device runtimes can be accurately characterized in just minutes.
    \item We propose two simple but very effective techniques to adapt MAPLE for diverse edge devices. We empirically show that our proposed techniques of targeted uniform sampling and performance counter normalization, have a significant impact on accuracy, resulting in average gains of up to +48.81\% compared to MAPLE on certain runtimes.
    \item We provide comprehensive experimental results illustrating how using only 10 measurements from an unseen edge runtime, the proposed MAPLE-Edge can achieve state-of-the-art accuracy results outperforming the state-of-the-art methods including MAPLE~\cite{abbasi2021maple} and HELP~\cite{lee2021hardware}.
\end{itemize}

The rest of the manuscript is organized as follows. Section~\ref{sec:method} describes the main methodology and the procedure of acquiring the training data from the edge devices. Section~\ref{sec:discussion} dives into the experimental results and compares the proposed method and competing algorithms. Finally we conclude the paper in Section~\ref{sec:conclusion}.

\section{Method}
\label{sec:method}
In this section, we describe the framework for predicting the latency on edge devices including the search space, hardware descriptor, hardware cost collection pipeline, preprocessing, and data augmentation strategies.

\subsection{MAPLE-Edge}
The seminal  MAPLE technique~\cite{abbasi2021maple} was evaluated on general purpose hardware including  Intel processors and server class GPUs. While the proposed algorithm outperformed other state-of-the-art techniques in estimating the latency, our experiments showed that MAPLE cannot always achieve strong results on ARM based embedded devices due to the added diversity in the device pool which needs further extension to address the issue. To this end, we propose a simple yet very effective trick to make the predictor generalize better: i) we take advantage of a targeted uniform sampling for selecting adaptation architectures, and ii) a new approach to represent hardware performance counters by the corresponding operator latency, to make the hardware descriptor adapt across diverse edge device runtimes.

\subsubsection{Targeted Uniform Adaptation Sampling}
The proposed MAPLE algorithm used a random selection approach to identify the adaptation set. While this approach makes the algorithm fairly easy and practical, it might result in instances where the selected samples may not be a good representation of the larger distribution.

To minimize this risk, MAPLE-Edge leverages a targeted uniform sampling strategy where adaptation samples are selected to maximize the probability of falling across the latency space. Specifically, by creating $N$ bins corresponding to the $N$ adaptation samples and ranking architectures based on the measured end-to-end latency for each device in the training pool, we improve the probability of selecting a diverse set of samples, thereby representing the space more effectively. The ranked architectures for each device are then distributed uniformly across the $N$ bins, yielding a total of $M \cdot \mathbb{X} / N$ architectures per bin, where $M$ is the number of training devices, and $\mathbb{X}$ is the number of architectures in the initial training set. One adaptation architecture is then randomly sampled per bin.

\begin{figure*}[h]
\begin{subfigure}{0.5\textwidth}
\includegraphics[width=1.0\linewidth, height=6cm]{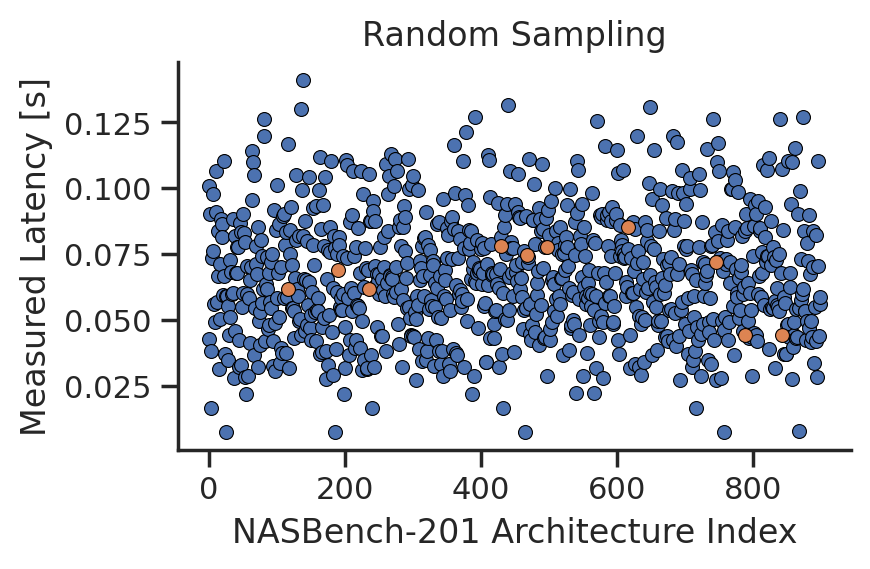}
\label{fig:random_sampling}
\end{subfigure}
\begin{subfigure}{0.5\textwidth}
\includegraphics[width=1.0\linewidth, height=6cm]{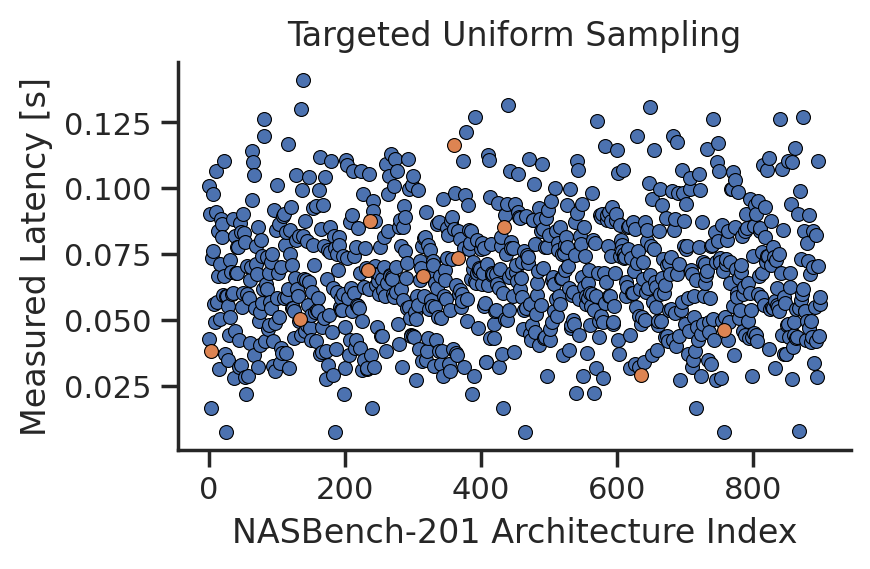}
\label{fig:targeted_uniform_sampling}
\end{subfigure}

\caption{Comparison of the random adaptation sampling technique used by MAPLE (left) versus the targeted uniform sampling strategy proposed by MAPLE-Edge (right). The orange dots highlight ten adaptation architectures that were sampled using both methods from the first 900 architectures in NAS-Bench-201. The targeted uniform sampler intentionally selects architectures from across the latency distribution whereas the random sampler has no prior, and may result in the selected architectures being drawn from a narrow band of the latency space. By maximizing coverage of the latency space, we ensure that the LPM is more likely to be trained on samples that are representative of the test device.}
\label{fig:random_vs_bucket_sampling}
\end{figure*}

The intuition behind this is that although the absolute latency values are different across device runtimes, the relative orderings of architectures tend to be similar. Thus, by merging the latency distributions of all architectures in the training pool together and sampling from the combined pool, we create a more robust prior to sample the adaptation architectures from. Figure~\ref{fig:random_vs_bucket_sampling} demonstrates the effect of the proposed Targeted Uniform sampling compared to random selection. As seen, the selected samples via the random selection exhibits similar latencies while the selected samples by the proposed sampling approach show a more diverse set spread across the distribution of training architectures.

\subsubsection{Performance Counter Normalization}
\label{sec:perf_counter_normalization}
Due to the limited diversity of devices in the training pool, MAPLE was able to effectively characterize devices using the raw performance counters obtained directly from the \textit{Perf} tool~\cite{perf_2022}. However on edge devices, there can be significant variance between the various runtimes and devices. Table~\ref{tab:mean_latencies} shows the mean latencies across the first 2700 neural network architectures in NAS-Bench-201 for an ImageNet style input of size $224 \times 224$ with batch size 1. The differences in latency between runtimes and even among devices of the same runtime imply that a hardware descriptor containing absolute performance counters measured over the entire duration of inference will be less suitable as a runtime descriptor, and as result the latency predictor will have to learn an additional mapping between an operator metric and its corresponding latency. Instead, to  provide more effective representation and accelerate learning in the predictor, MAPLE-Edge normalizes all performance counters collected by \textit{Perf} by the measured latency of the respective operator. By dividing each performance counter by the latency, this effectively transforms the hardware descriptor from an embedding of absolute values (e.g. cache-misses) to an embedding of relative rates (e.g. cache-misses per second).

\begin{table}[]
\caption{Average measured latency across architectures [0,2699] in NAS-Bench-201 (averaged over 50 trials each), using TensorFlow Lite and TensorRT runtimes. The large variance in latency between the different runtimes makes it difficult for the MAPLE hardware descriptor to effectively represent diverse device runtimes. We mitigate this issue by applying the performance counter normalization trick described in Section~\ref{sec:perf_counter_normalization}}
\label{tab:mean_latencies}
\begin{tabular}{@{}l|c|c@{}}
\toprule
                        & \multicolumn{2}{c}{\textbf{Runtime}}                         \\ \midrule
\textbf{Device}         & \textbf{TensorFlow Lite {[}s{]}} & \textbf{TensorRT {[}s{]}} \\ \midrule \midrule
\textbf{Jetson TX1}     & 1.0 $\pm$ 0.5                        & 0.05 $\pm$ 0.01               \\
\textbf{Jetson TX2}     & 0.9 $\pm$ 0.5                        & 0.03 $\pm$ 0.01             \\
\textbf{Jetson Nano}    & 1.1 $\pm$ 0.6                        & 0.07 $\pm$ 0.02               \\
\textbf{Raspberry Pi 4B} & 0.6 $\pm$ 0.2                        & -                         \\ \bottomrule
\end{tabular}
\end{table}

\subsection{Data Collection Pipeline}
Embedded vision applications routinely operate under tight deadlines with low latency constraints. To maximize performance in production, expert domain knowledge can be used to optimize deep learning models to more efficient formats (e.g. through the use of runtime specific compilers), as well as hardware modifications (e.g. overclocking, memory swapping, power usage maximization). Similar to HW-NAS-Bench~\cite{li2021hw}, MAPLE-Edge leverages this domain knowledge to build an optimized, generic hardware-cost collection pipeline that automates the process of collecting latency measurements (as seen in Figure~\ref{fig:data_collection_diagram}).

On the NVIDIA Jetson devices, MAPLE-Edge executes inference on the GPU using models compiled to serialized TensorRT (TRT) engine format, while CPU based inference is executed on all devices using the TensorFlow Lite (TFLite) runtime engine. Unlike HW-NAS-Bench however, MAPLE-Edge also records hardware performance counters for all 15 operations in NAS-Bench-201~\cite{dong2020bench}, by executing the Linux \textit{Perf}~\cite{perf_2022} tool while running inference.

\subsection{Edge Dataset}
Similar to MAPLE~\cite{abbasi2021maple}, BRP-NAS~\cite{dudziak2020brp}, HELP~\cite{lee2021hardware}, and HW-NAS-Bench~\cite{li2021hw}, MAPLE-Edge also uses the NAS-BENCH-201 \cite{dong2020bench} dataset for all experiments. NAS-BENCH-201 is a collection of 15,625 neural cell candidates with each architecture having a fixed cell topology with five possible operations in its search space including \{none, skip-connection, conv1x1, conv3x3, avgpool3x3\}. Each operation allows 16, 32, or 64 input and output channels, yielding a total of 15 possible variations for the operations that can be used in designing new architectures.

The key observation in MAPLE was that the latency search space can be effectively characterized at the operator level, if the operator latency is augmented with a hardware description vector. Unlike methods like nn-Meter, MAPLE quickly builds a hardware descriptor for a device or runtime by executing the Linux tool \textit{Perf} to measure 10 hardware performance counters while a model gets executed.

MAPLE-Edge follows a similar strategy, but builds a hardware descriptor using only 6 out of the original 10 performance counters that were found to be supported by default on all Linux kernels. These include performance counters for CPU-cycles, instructions, cache-references, cache-misses, level one (L1) data cache loads, and L1 data cache load misses. Even though we do not collect any metrics relating to LLC (last level cache) counters due to limited support on some ARM based processors, we show that the results are still competitive and achieve strong gains over all baselines. Each operation is characterized by running inference for $N=1000$ runs, and the \textit{Perf} tool is launched from a Python Subprocess to profile the execution of the inference occurring in the main thread.

To create our dataset, we measure the end-to-end latency of the first 2,700 architectures in NAS-Bench-201 on a variety of different edge devices and runtimes. These devices were selected due to their ubiquity in industrial embedded vision applications and include a i) 4GB 4xA57 core Jetson Nano with 128 NVIDIA Maxwell CUDA cores, ii) 4GB 4xA57 core Jetson TX1 with 256 NVIDIA Maxwell CUDA cores, iii) 8GB 6xA57 core Jetson TX2 with 256 NVIDIA Pascal CUDA cores, and iv) a 4GB 4xA72 core Raspberry Pi 4B running 64-bit Raspbian OS. Of these 2,700 architectures, we sample from the first 900 architectures for training our predictor, and report the test accuracy of all methods on architectures in range [1,800, 2,699]. This test strategy lets us evaluate how well the method performs on completely unseen architectures.

To train the regression model, we collect an initial set of architectures $\mathbb{X}$ from all devices in the training pool, along with an adaptation set of few (i.e., typically 10) architectures $\hat{\mathbb{X}}$ from devices in the test pool. The initial set $\mathbb{X}$ helps the model generalize to the search space while the adaptation samples $\hat{\mathbb{X}}$ help the predictor specialize to the test devices.

\subsection{Data Augmentation}
MAPLE-Edge also applies a simple yet effective data augmentation strategy where all selected adaptation samples from the test device are naively cloned $K-1$ times to increase their representation in the training pool. Given the limited number of samples we collect, we empirically find that this approach helps the model prioritize the adaptation samples. However, we find that this data augmentation technique may result in worse performance if the adaptation samples are poorly chosen to begin with (such as with a random selection strategy) instead of a more representative selection strategy such as targeted uniform sampling, which is more robust to outliers. Unless otherwise noted, all experiments in this paper use an augmentation factor of $K=7$, meaning that for each adaptation sample, 6 additional clones are created.

We follow the same training strategy proposed in  MAPLE~\cite{abbasi2021maple}, including the feedforward regression model, model training regime, loss function, and evaluation criteria. As such, training MAPLE-Edge remains fast, and takes less than 2 mins to train from scratch until completion (200 epochs, batch size 128) on a single NVIDIA GTX-1080Ti machine.

\begin{table*}[]
\begin{center}
\caption{Summary of results comparing MAPLE-Edge to a LUT based estimator as well as HELP, and MAPLE baselines. The method column describes the method and training paradigm used. For example, \textit{MAPLE-Edge TFLite} implies that each test device in that row was trained on MAPLE-Edge using a device pool containing only TensorFlow Lite runtimes. All methods were trained for ten trials with unique set adaptation samples (for each trial) used across all methods for consistency (selected using our targeted uniform sampling technique).}
\label{tab:results_summary}
\setlength\tabcolsep{2pt}
\begin{tabular}{@{}l|c|c|c|c||c|c|c@{}}
\toprule
 &
  \multicolumn{7}{c}{\textbf{Test Runtime - Mean $\boldsymbol{\pm10\%}$ Accuracy}} \\ \midrule
\textbf{Method} &
  \textbf{\begin{tabular}[c]{@{}c@{}}Raspberry Pi\\ TFLite\end{tabular}} &
  \textbf{\begin{tabular}[c]{@{}c@{}}Jetson TX1\\ TFLite\end{tabular}} &
  \textbf{\begin{tabular}[c]{@{}c@{}}Jetson TX2\\ TFLite\end{tabular}} &
  \textbf{\begin{tabular}[c]{@{}c@{}}Jetson Nano\\ TFLite\end{tabular}} &
  \textbf{\begin{tabular}[c]{@{}c@{}}Jetson TX1\\ TRT\end{tabular}} &
  \textbf{\begin{tabular}[c]{@{}c@{}}Jetson TX2\\ TRT\end{tabular}} &
  \textbf{\begin{tabular}[c]{@{}c@{}}Jetson Nano\\ TRT\end{tabular}} \\ \midrule \midrule
\small LUT &
  \textbf{55.18} &
  80.96 &
  80.22 &
  79.96 &
  12.07 &
  0.00 &
  7.85 \\ \midrule \midrule
\begin{tabular}[c]{@{}l@{}} \small HELP  TFLite\end{tabular} &
  42.77$\pm$5.07 &
  54.90$\pm$9.62 &
  58.82$\pm$12.00 &
  58.67$\pm$11.99 &
  - &
  - &
  - \\
\begin{tabular}[c]{@{}l@{}}  \small MAPLE  TFLite\end{tabular} &
  45.04$\pm$7.69 &
  \textbf{99.22$\pm$0.31} &
  96.81$\pm$1.75 &
  \textbf{98.53$\pm$0.83} &
  - &
  - &
  - \\
\begin{tabular}[c]{@{}l@{}}\small MAPLE-Edge  TFLite\end{tabular} &
  \textbf{54.00$\pm$11.43} &
  98.89$\pm$0.35 &
  \textbf{97.4$\pm$1.59} &
  96.18$\pm$1.85 &
  - &
  - &
  - \\ \midrule \midrule
\begin{tabular}[c]{@{}l@{}}\small HELP  TRT\end{tabular} &
  - &
  - &
  - &
  - &
  79.88$\pm$6.34 &
  82.64$\pm$5.25 &
  79.54$\pm$5.36 \\
\begin{tabular}[c]{@{}l@{}}\small MAPLE TRT\end{tabular} &
  - &
  - &
  - &
  - &
  96.52$\pm$1.79 &
  \textbf{96.64$\pm$2.05} &
  \textbf{94.37$\pm$2.76} \\
\begin{tabular}[c]{@{}l@{}}\small MAPLE-Edge TRT\end{tabular} &
  - &
  - &
  - &
  - &
  \textbf{96.68$\pm$1.01} &
  94.89$\pm$2.37 &
  93.67$\pm$3.07 \\ \midrule \midrule
\begin{tabular}[c]{@{}l@{}}\small HELP TRT + TFLite\end{tabular} &
  47.62$\pm$5.60 &
  52.28$\pm$9.55 &
  53.43$\pm$10.45 &
  57.43$\pm$10.43 &
  67.93$\pm$4.80 &
  63.56$\pm$7.99 &
  68.50$\pm$8.36 \\
\begin{tabular}[c]{@{}l@{}}\small MAPLE TRT + TFLite\end{tabular} &
  51.37$\pm$3.28 &
  98.94$\pm$0.67 &
  96.89$\pm$1.93 &
  \textbf{98.40$\pm$0.44} &
  45.61$\pm$23.69 &
  27.59$\pm$20.04 &
  80.01$\pm$8.13 \\
\begin{tabular}[c]{@{}l@{}}\small MAPLE-Edge TRT + TFLite\end{tabular} &
  \textbf{54.23$\pm$4.21} &
  \textbf{99.34$\pm$0.24} &
  \textbf{99.19$\pm$0.42} &
  98.19$\pm$0.94 &
  \textbf{91.71$\pm$2.00} &
  \textbf{76.64$\pm$6.66} &
  \textbf{82.08$\pm$7.05} \\ \bottomrule
\end{tabular}
\end{center}
\end{table*}

\section{Results \& Discussion}
\label{sec:discussion}
In this section we evaluate the performance of the proposed MAPLE-Edge approach and compared it with the state-of-the-art method in predicting the latency of neural network architectures on different edge devices. We also study the effect of each proposed techniques individually to illustrate the importance of the each step in improving the model accuracy.

\subsection{Experimental Setup}
To evaluate the performance of the proposed MAPLE-Edge, we compare it against look-up table (LUT) based method as the baseline and the state-of-the-art algorithms including HELP~\cite{lee2021hardware} and MAPLE~\cite{abbasi2021maple}. For all experiments, HELP, MAPLE, and MAPLE-Edge were trained using architectures [0,899] from NAS-BENCH-201 across 10 trials, each with a unique adaptation set generated using our uniform targeted sampling technique. The 10 adaptation samples were kept consistent across all methods to ensure that the results are due to the technique's ability to generalize and not because the adaptation samples are drawn from a more representative distribution. A more detailed study comparing the original MAPLE algorithm with each of our proposed improvements can be found in Section~\ref{sec:maple_improvements} and Table~\ref{tab:maple_improvements}.
Similar to BRP-NAS and MAPLE, we evaluate all methods using a $\pm 10$\% error-bound accuracy metric, which describes the percentage of models with predicted latency within the corresponding error bound relative to the measured latency \cite{dudziak2020brp}. All methods were also evaluated on architectures [1800,2699] for consistency.

\subsection{Results}
\begin{figure*}
  \centering
  \includegraphics[width=1.0\linewidth]{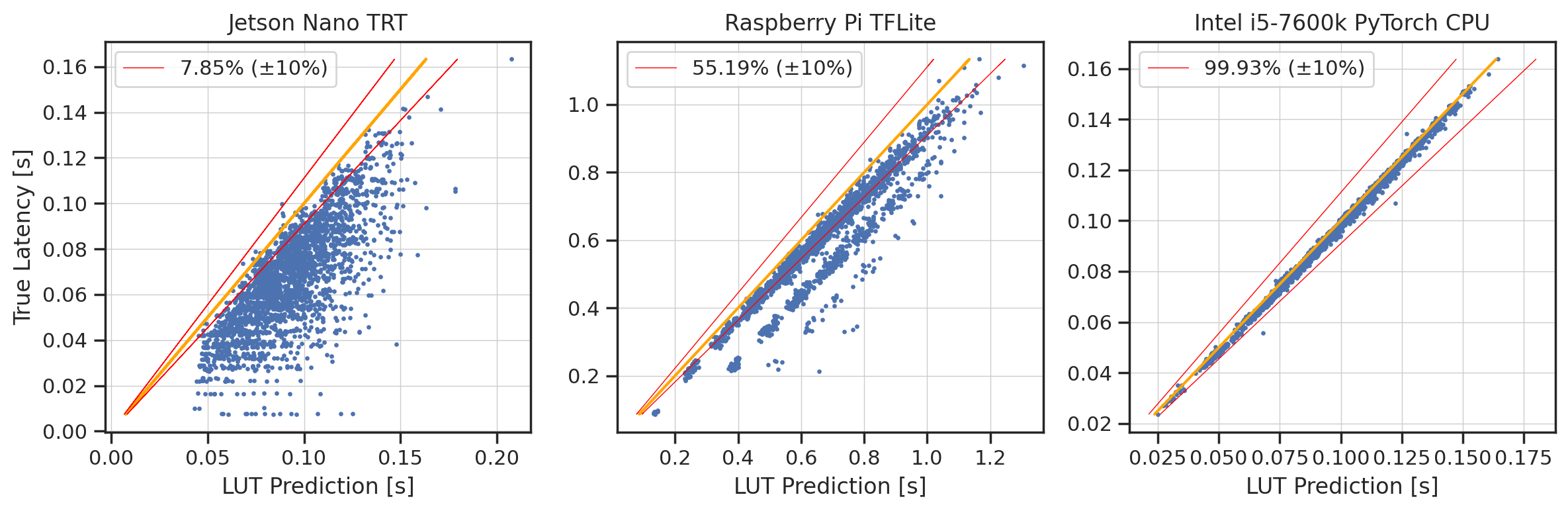}

  \caption{Look-up-table (LUT) latency predictions versus true latency for edge runtimes Jetson Nano TRT, and Raspberry Pi TFLite compared to non-edge runtime (PyTorch CPU inference on Intel i5-7600k). The red lines represent the $\pm10\%$ error bound and the yellow line represents a perfect prediction. Note how LUT based estimation tends to overestimate the latency for inference executed on optimized graph, as it cannot take into account the runtime optimizations performed by the execution engine.}
  \label{fig:lut_baseline}
\end{figure*}

Table~\ref{tab:results_summary} shows the comprehensive comparison analysis of the proposed method and competing algorithms. We can see how the LUT baseline performs well on the TensorFlow Lite runtimes but fails to generalize to any of the TensorRT runtimes. This is because the LUT approach estimates the total end-to-end latency by adding together the latency of each individual block and operator, without accounting for any runtime optimizations made by the execution engine. For example, NVIDIA's TensorRT automatically eliminates layers with unused outputs and fuses together distinct layers to improve efficiency of running networks on the GPU~\cite{nvidia_tensorrt_2022}. Because of such optimizations, the resulting graph may not bear a lot of semblance to the original stack of layers, thus rendering the LUT based approach incapable of accurately estimating architecture latency. This can be easily seen in Figure~\ref{fig:lut_baseline} which shows how skewed the latency distribution looks for optimized edge runtimes, due to the LUT predictor overestimating architecture latency. On the other hand for  un-optimized runtimes such as PyTorch under CPU based execution on an Intel processor, we can see how LUT based estimators achieve nearly perfect accuracy. This is why LUT based approaches have continued to be successful in a variety of non-embedded contexts, but this further underscores the need for more robust latency estimation methods for optimized edge runtimes.

The performance of all other methods were evaluated using two different types of training pools; runtime based pooling and combined pooling. In runtime based pooling, the initial set of samples are drawn from devices sharing the same runtime as the test device, excluding the test device itself. For the combined pooling setup, the initial set of samples are drawn from all available devices, excluding the test device. In almost all cases, we can see in Table~\ref{tab:results_summary} that MAPLE-Edge shows strong gains over the HELP baseline, especially in the case of runtime-based pooling where fewer devices are available. This is particularly significant in an embedded vision context as most applications would only have the infrastructure to support a single runtime instead of all available runtimes. By being able to effectively generalize to an unseen device on the same runtime, MAPLE-Edge can propel significant advancements in the field of neural architecture search for embedded vision contexts.

The effectiveness of MAPLE-Edge is most evident in Table~\ref{tab:results_single_device} where we attempt to predict the latency on a test device, using only a single device in the training pool. The reported results are obtained using leave-one-out cross-validation, where we rotate a device into the training pool to ensure that the results are representative of all available devices for that particular runtime. Here we can see that despite only being trained on 900 samples from a single device, MAPLE-Edge can effectively estimate the latency of architectures being executed on an unseen device. The strong gains over both HELP and MAPLE shows that the proposed LPM algorithm is very capable of estimating latency on optimized graphs, something that current state-of-the-art estimators have trouble with.

\subsection{Improvements Over MAPLE}
\label{sec:maple_improvements}
Table~\ref{tab:maple_improvements} provide a detailed glimpse into the impact of our proposed improvements on the MAPLE algorithm. The results labelled with a $\star$ represent the performance of original MAPLE (i.e. with no performance counter normalization, no data augmentation, and using a random adaptation sample selection strategy). Although MAPLE shows strong performance across the board (90+\% top $\pm10\%$ accuracy), it appears to get particularly challenged when presented with a diverse group of devices such as the device pool with combined TRT and TFLite runtimes. In such a scenario, the regression model appears unable to effectively generalize the hardware descriptor it has learned, to the adaptation samples presented from the test device. This is particularly visible in the Jetson TX1-TRT and Jetson TX2-TRT columns for MAPLE (TRT+TFLite) with baseline accuracies near 47\% and 27\% respectively. By applying the targeted uniform adaptation sampling strategy, the accuracy is seen to increase $+1.99\%$ on average. Stacking this with the latency normalization provides an average gain of $+7.24\%$ as well as a significant jump of $+28.93\%$ and $+48.81\%$ for the lowest performing TX1-TRT and TX2-TRT runtimes. Thus by only applying these small tricks, we show that the core ideas behind MAPLE can be successfully extended to diverse hardware pools as well, and that MAPLE-Edge can be used in Neural Architecture Search as an effective LPM tailored towards edge devices.

\begin{table}
\centering
\caption{Results from predicting the latency on each test device, using only one device in the training pool. Reported values are mean $\pm10\%$ accuracy, obtained using leave-one-out cross validation across multiple training devices.}
\label{tab:results_single_device}
\setlength\tabcolsep{2.5pt}
\begin{tabular}{@{}l|ccc@{}}
\toprule                                                        & \multicolumn{3}{c}{\textbf{Test Runtime - Mean $\boldsymbol{\pm10\%}$ Accuracy}}        \\ \midrule
\textbf{Method} &
  \textbf{\begin{tabular}[c]{@{}c@{}}Jetson TX1\\ TRT\end{tabular}} &
  \textbf{\begin{tabular}[c]{@{}c@{}}Jetson TX2\\ TRT\end{tabular}} &
  \textbf{\begin{tabular}[c]{@{}c@{}}Jetson Nano\\ TRT\end{tabular}} \\ \midrule \midrule
\begin{tabular}[c]{@{}l@{}}\small HELP TRT\end{tabular}      & 78.95$\pm$5.02          & 76.35$\pm$7.13          & 76.44$\pm$6.54           \\ \midrule
\begin{tabular}[c]{@{}l@{}}\small MAPLE TRT\end{tabular} &
  \multicolumn{1}{l}{76.63$\pm$19.05} &
  \multicolumn{1}{l}{81.93$\pm$12.53} &
  \multicolumn{1}{l}{59.52$\pm$26.39} \\ \midrule
\begin{tabular}[c]{@{}l@{}}\small MAPLE-Edge TRT\end{tabular} & \textbf{95.23$\pm$3.27} & \textbf{96.59$\pm$2.09} & \textbf{91.04$\pm$4.567}
\end{tabular}
\end{table}

\begin{table*}[]
\begin{center}
\caption{Detailed results showing how the additions we propose to MAPLE (targeted uniform adaptation sampling, performance counter normalization, and data augmentation) on average provide additive gains over the baseline method. The method column is broken down into 3 main sections: (top) These are the results from experiments where the device pool only consists of TensorFlow Lite runtimes. (middle) These are the results from experiments where the device pool only consists of TensorRT runtimes. (bottom) These are the results from experiments where the device pool consists of both TensorRT and TensorFlow Lite runtimes. The top row in each section is labelled with a $\star$ signifying that these are the absolute accuracy values for the original MAPLE algorithm when evaluated on the corresponding test device runtime. A $\dagger$, $\oplus$, and $\cup$ labels correspond respectively to targeted uniform adaptation sampling, latency based performance counter normalization, and data augmentation. These techniques are shown to augment the accuracy of MAPLE, with each technique providing additional additive gains over the baseline on average.}
\label{tab:maple_improvements}
\setlength\tabcolsep{1pt}
\begin{tabular}{@{}l|c|c|c|c||c|c|c@{}}
\toprule
 &
  \multicolumn{7}{c}{\textbf{Test Runtime - Mean $\boldsymbol{\pm10\%}$ Accuracy}} \\ \midrule
\textbf{Method} &
  \textbf{\begin{tabular}[c]{@{}c@{}}Raspberry \\Pi TFLite\end{tabular}} &
  \textbf{\begin{tabular}[c]{@{}c@{}}Jetson TX1\\ TFLite\end{tabular}} &
  \textbf{\begin{tabular}[c]{@{}c@{}}Jetson TX2\\ TFLite\end{tabular}} &
  \textbf{\begin{tabular}[c]{@{}c@{}}Jetson Nano\\ TFLite\end{tabular}} &
  \textbf{\begin{tabular}[c]{@{}c@{}}Jetson TX1\\ TRT\end{tabular}} &
  \textbf{\begin{tabular}[c]{@{}c@{}}Jetson TX2\\ TRT\end{tabular}} &
  \textbf{\begin{tabular}[c]{@{}c@{}}Jetson Nano\\ TRT\end{tabular}} \\ \midrule \midrule
\begin{tabular}[c]{@{}l@{}}  \small MAPLE  (TFLite) $\star$\\\end{tabular} &
  41.70 &
  98.53 &
  94.39 &
  95.33 &
  - &
  - &
  - \\
\begin{tabular}[c]{@{}l@{}}  \small MAPLE  (TFLite) $\star \dagger$\\\end{tabular} &
  +3.34 &
  +0.69 &
  +2.42 &
  \textbf{+3.20} &
  - &
  - &
  - \\
\begin{tabular}[c]{@{}l@{}}  \small MAPLE (TFLite) $\star \dagger \oplus$ \\\end{tabular} &
  +3.09 &
  \textbf{+0.83} &
  \textbf{+3.73} &
  +2.18 &
  - &
  - &
  - \\
\begin{tabular}[c]{@{}l@{}}\small MAPLE-Edge  (TFLite) $\star \dagger \oplus \cup$ \end{tabular} &
  \textbf{+12.30} &
  +0.36 &
  +3.01 &
  +0.85 &
  - &
  - &
  - \\ \midrule \midrule
\begin{tabular}[c]{@{}l@{}}  \small MAPLE  (TRT) $\star$\\\end{tabular} &
  - &
  - &
  - &
  - &
  94.10 &
  94.66 &
  91.91 \\
\begin{tabular}[c]{@{}l@{}}  \small MAPLE  (TRT) $\star \dagger$\\\end{tabular} &
  - &
  - &
  - &
  - &
  +2.42 &
  \textbf{+1.98} &
  +2.46 \\
\begin{tabular}[c]{@{}l@{}}  \small MAPLE (TRT) $\star \dagger \oplus$ \\\end{tabular} &
  - &
  - &
  - &
  - &
  \textbf{+3.32} &
  -1.74 &
  -0.81 \\
\begin{tabular}[c]{@{}l@{}}\small MAPLE-Edge  (TRT) $\star \dagger \oplus \cup$ \end{tabular} &
  - &
  - &
  - &
  - &
  +2.58 &
  +0.23 &
  \textbf{+1.76} \\ \midrule \midrule
\begin{tabular}[c]{@{}l@{}}  \small MAPLE  (TRT+TFLite) $\star$\\\end{tabular} &
  44.86 &
  98.96 &
  93.99 &
  96.59 &
  46.89 &
  27.09 &
  79.06 \\
\begin{tabular}[c]{@{}l@{}}  \small MAPLE  (TRT+TFLite) $\star \dagger$\\\end{tabular} &
  +6.51 &
  -0.02 &
  +2.90 &
  \textbf{+1.81} &
  -1.28 &
  +0.50 &
  +0.95 \\
\begin{tabular}[c]{@{}l@{}}  \small MAPLE (TRT+TFLite) $\star \dagger \oplus$ \\\end{tabular} &
  +4.48 &
  \textbf{+0.42} &
  +5.08 &
  +1.18 &
  +28.93 &
  +48.81 &
  +1.83 \\
\begin{tabular}[c]{@{}l@{}}\small MAPLE-Edge  (TRT+TFLite) $\star \dagger \oplus \cup$ \end{tabular} &
  \textbf{+9.37} &
  +0.38 &
  \textbf{+5.20} &
  +1.60 &
  \textbf{+44.82} &
  \textbf{+49.55} &
  \textbf{+3.02} \\ \bottomrule
\end{tabular}
\end{center}
\end{table*}

\subsection{Effect of Adaptation Samples}

\begin{figure}[t]
  \centering
   \includegraphics[width=1.0\linewidth]{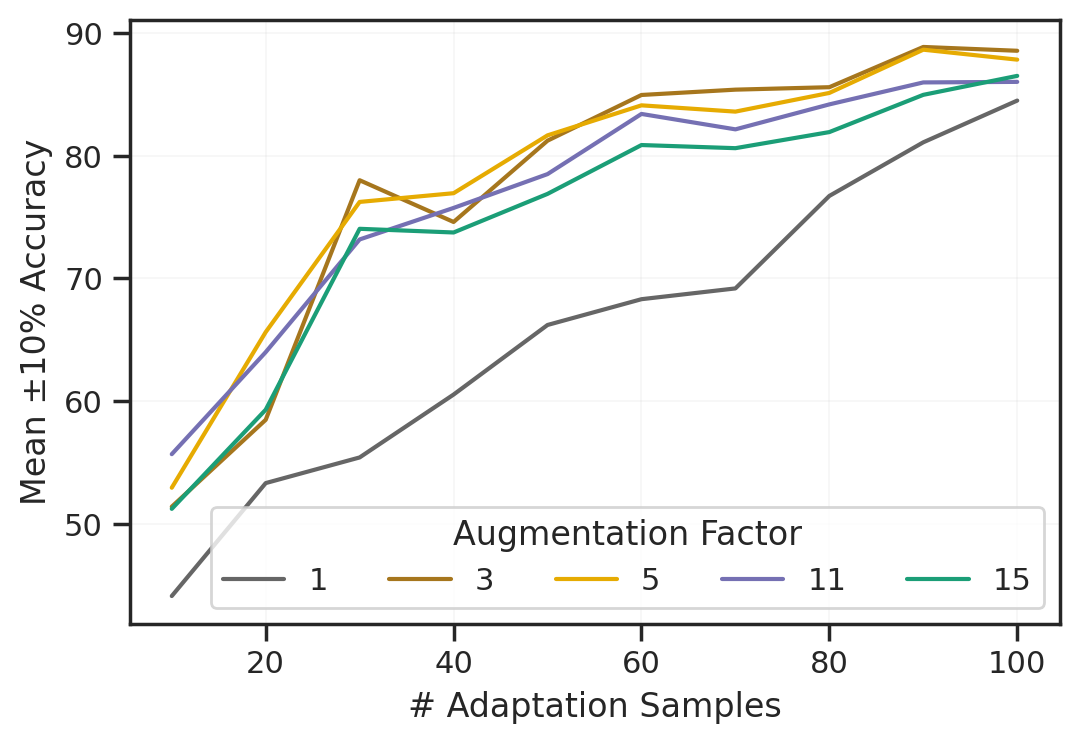}

   \caption{Impact of adaptation samples and augmentation factor on mean $\pm$10\% accuracy for Raspberry Pi TensorFlow Lite runtime. Each curve shows the result for using different number of augmentation sample $K$. Note, how increasing the number of adaptation samples from 10 to 100 provides a nearly 40\% boost in prediction accuracy when no data augmentation is applied. On the other hand, we see that data augmentation proves most beneficial when adaptation samples are limited, and that beyond a certain point, additional data augmentation may in fact even harm accuracy. The training pool for this experiment consisted of all edge device runtimes excluding Raspberry Pi TFLite.}
   \label{fig:num_adaptations}
\end{figure}

Despite showing strong gains against baselines for all runtimes on the Jetson family of devices, MAPLE-Edge does not make any gains against the LUT approach for the Raspberry-Pi TensorFlow Lite runtime. To investigate which hyperparameters affect our method the most, we conduct an ablation study and experiment with varying the values of the number of adaptation samples (previously fixed at 10) as well as the number of clones created per adaptation sample. The results of our experiments can be seen in Figure~\ref{fig:num_adaptations} which shows how additional adaptation samples and augmentation factor work together to significantly improve generalization, boosting the estimation accuracy on the Raspberry Pi TensorFlow Lite runtime from around 50\% accuracy to nearly 90\% accuracy. Although this requires almost 10x the number of adaptation samples, we believe that the additional time spent collecting samples (10 minutes vs. 100 minutes) is a tradeoff worth making in industrial scenarios, as it can significantly improve the latency estimation accuracy on embedded devices.

\section{Conclusion}
\label{sec:conclusion}
In this work, we propose MAPLE-Edge, a simple yet highly effective extension to MAPLE, enabling the most accurate latency estimation on embedded devices including optimized graphs. MAPLE-Edge leverages a reduced hardware descriptor that characterizes the device using just 6 performance counters that are available by default on all Linux kernels. We show that by measuring CPU based performance counters while executing inference using optimized graphs on embedded devices, we can more accurately estimate latency on TensorRT and TensorFlow Lite runtimes, compared to previous state-of-the-art approaches like HELP. We also observe additional gains over MAPLE by applying our proposed targeted uniform sampling and normalization strategies. We validate the proposed methods by conducting experiments with ten random trials each, using a unique set of adaptation samples in each trial. We find that on average, with ten adaptation samples and a targeted uniform adaptation sampling technique, MAPLE-Edge yields an improvement of 26.08\% over HELP and 7.65\% increase over MAPLE when there is more than one device in the training pool. MAPLE-Edge exhibits particularly strong gains when estimating the latency when trained on just 900 architectures from a single device, yielding 17.04\% improvements over HELP and 21.59\% improvement over MAPLE on average. By requiring significantly fewer adaptation samples than other techniques, MAPLE-Edge demonstrates that it is possible to effectively estimate latency on optimized edge runtimes, paving the way for more efficient architectures to be discovered for embedded vision using Neural Architecture Search.

%\clearpage

{\small
\bibliographystyle{ieee_fullname}
\bibliography{egbib}

\begin{thebibliography}{10}\itemsep=-1pt

\bibitem{abbasi2021maple}
Saad Abbasi, Alexander Wong, and Mohammad~Javad Shafiee.
\newblock Maple: Microprocessor a priori for latency estimation.
\newblock In {\em Proceedings of the IEEE Conference on Computer Vision and
  Pattern Recognition}, 2022.

\bibitem{cai2019once}
Han Cai, Chuang Gan, Tianzhe Wang, Zhekai Zhang, and Song Han.
\newblock Once-for-all: Train one network and specialize it for efficient
  deployment.
\newblock {\em arXiv preprint arXiv:1908.09791}, 2019.

\bibitem{cai2018proxylessnas}
Han Cai, Ligeng Zhu, and Song Han.
\newblock Proxylessnas: Direct neural architecture search on target task and
  hardware.
\newblock {\em arXiv preprint arXiv:1812.00332}, 2018.

\bibitem{nvidia_tensorrt_2022}
NVIDIA Corporation.
\newblock Nvidia tensorrt, 2022.

\bibitem{dai2019chamnet}
Xiaoliang Dai, Peizhao Zhang, Bichen Wu, Hongxu Yin, Fei Sun, Yanghan Wang,
  Marat Dukhan, Yunqing Hu, Yiming Wu, Yangqing Jia, et~al.
\newblock Chamnet: Towards efficient network design through platform-aware
  model adaptation.
\newblock In {\em Proceedings of the IEEE Conference on Computer Vision and
  Pattern Recognition}, pages 11398--11407, 2019.

\bibitem{dong2020bench}
Xuanyi Dong and Yi Yang.
\newblock Nas-bench-201: Extending the scope of reproducible neural
  architecture search.
\newblock {\em arXiv preprint arXiv:2001.00326}, 2020.

\bibitem{dudziak2020brp}
Lukasz Dudziak, Thomas Chau, Mohamed Abdelfattah, Royson Lee, Hyeji Kim, and
  Nicholas Lane.
\newblock Brp-nas: Prediction-based nas using gcns.
\newblock {\em Advances in Neural Information Processing Systems},
  33:10480--10490, 2020.

\bibitem{he2017mask}
Kaiming He, Georgia Gkioxari, Piotr Doll{\'a}r, and Ross Girshick.
\newblock Mask r-cnn.
\newblock In {\em Proceedings of the IEEE Conference on Computer Vision and
  Pattern Recognition}, pages 2961--2969, 2017.

\bibitem{he2016deep}
Kaiming He, Xiangyu Zhang, Shaoqing Ren, and Jian Sun.
\newblock Deep residual learning for image recognition.
\newblock In {\em Proceedings of the IEEE Conference on Computer Vision and
  Pattern Recognition}, pages 770--778, 2016.

\bibitem{lee2021hardware}
Hayeon Lee, Sewoong Lee, Song Chong, and Sung~Ju Hwang.
\newblock Hardware-adaptive efficient latency prediction for nas via
  meta-learning.
\newblock {\em Advances in Neural Information Processing Systems}, 34, 2021.

\bibitem{li2021hw}
Chaojian Li, Zhongzhi Yu, Yonggan Fu, Yongan Zhang, Yang Zhao, Haoran You,
  Qixuan Yu, Yue Wang, and Yingyan Lin.
\newblock Hw-nas-bench: Hardware-aware neural architecture search benchmark.
\newblock {\em arXiv preprint arXiv:2103.10584}, 2021.

\bibitem{liu2018darts}
Hanxiao Liu, Karen Simonyan, and Yiming Yang.
\newblock Darts: Differentiable architecture search.
\newblock {\em arXiv preprint arXiv:1806.09055}, 2018.

\bibitem{perf_2022}
The Linux~Kernel Organization.
\newblock perf: Linux profiling with performance counters, 2022.

\bibitem{pham2018efficient}
Hieu Pham, Melody Guan, Barret Zoph, Quoc Le, and Jeff Dean.
\newblock Efficient neural architecture search via parameters sharing.
\newblock In {\em International Conference on Machine Learning}, pages
  4095--4104. PMLR, 2018.

\bibitem{redmon2018yolov3}
Joseph Redmon and Ali Farhadi.
\newblock Yolov3: An incremental improvement.
\newblock {\em arXiv preprint arXiv:1804.02767}, 2018.

\bibitem{ren2015faster}
Shaoqing Ren, Kaiming He, Ross Girshick, and Jian Sun.
\newblock Faster r-cnn: Towards real-time object detection with region proposal
  networks.
\newblock {\em Advances in Neural Information Processing Systems}, 28, 2015.

\bibitem{ronneberger2015u}
Olaf Ronneberger, Philipp Fischer, and Thomas Brox.
\newblock U-net: Convolutional networks for biomedical image segmentation.
\newblock In {\em International Conference on Medical image computing and
  computer-assisted intervention}, pages 234--241. Springer, 2015.

\bibitem{simonyan2014very}
Karen Simonyan and Andrew Zisserman.
\newblock Very deep convolutional networks for large-scale image recognition.
\newblock {\em arXiv preprint arXiv:1409.1556}, 2014.

\bibitem{tan2019mnasnet}
Mingxing Tan, Bo Chen, Ruoming Pang, Vijay Vasudevan, Mark Sandler, Andrew
  Howard, and Quoc~V Le.
\newblock Mnasnet: Platform-aware neural architecture search for mobile.
\newblock In {\em Proceedings of the IEEE Conference on Computer Vision and
  Pattern Recognition}, pages 2820--2828, 2019.

\bibitem{tan2019efficientnet}
Mingxing Tan and Quoc Le.
\newblock Efficientnet: Rethinking model scaling for convolutional neural
  networks.
\newblock In {\em International Conference on Machine Learning}, pages
  6105--6114. PMLR, 2019.

\bibitem{wan2020fbnetv2}
Alvin Wan, Xiaoliang Dai, Peizhao Zhang, Zijian He, Yuandong Tian, Saining Xie,
  Bichen Wu, Matthew Yu, Tao Xu, Kan Chen, et~al.
\newblock Fbnetv2: Differentiable neural architecture search for spatial and
  channel dimensions.
\newblock In {\em Proceedings of the IEEE Conference on Computer Vision and
  Pattern Recognition}, pages 12965--12974, 2020.

\bibitem{wang2020hat}
Hanrui Wang, Zhanghao Wu, Zhijian Liu, Han Cai, Ligeng Zhu, Chuang Gan, and
  Song Han.
\newblock Hat: Hardware-aware transformers for efficient natural language
  processing.
\newblock {\em arXiv preprint arXiv:2005.14187}, 2020.

\bibitem{wu2019fbnet}
Bichen Wu, Xiaoliang Dai, Peizhao Zhang, Yanghan Wang, Fei Sun, Yiming Wu,
  Yuandong Tian, Peter Vajda, Yangqing Jia, and Kurt Keutzer.
\newblock Fbnet: Hardware-aware efficient convnet design via differentiable
  neural architecture search.
\newblock In {\em Proceedings of the IEEE Conference on Computer Vision and
  Pattern Recognition}, pages 10734--10742, 2019.

\bibitem{xu2020latency}
Yuhui Xu, Lingxi Xie, Xiaopeng Zhang, Xin Chen, Bowen Shi, Qi Tian, and Hongkai
  Xiong.
\newblock Latency-aware differentiable neural architecture search.
\newblock {\em arXiv preprint arXiv:2001.06392}, 2020.

\bibitem{zhang2021nn}
Li~Lyna Zhang, Shihao Han, Jianyu Wei, Ningxin Zheng, Ting Cao, Yuqing Yang,
  and Yunxin Liu.
\newblock Nn-meter: Towards accurate latency prediction of deep-learning model
  inference on diverse edge devices.
\newblock In {\em Proceedings of the 19th Annual International Conference on
  Mobile Systems, Applications, and Services}, pages 81--93, 2021.

\bibitem{zoph2016neural}
Barret Zoph and Quoc~V Le.
\newblock Neural architecture search with reinforcement learning.
\newblock {\em arXiv preprint arXiv:1611.01578}, 2016.

\end{thebibliography}
}

\end{document}